\documentclass[a4paper,3p,12pt]{elsarticle}

\usepackage{fontspec}
\usepackage{etoolbox}
\makeatletter
\patchcmd{\emailauthor}{\ttfamily}{\rmfamily}{}{
  \PackageWarning{preamble}{Unable to harmonize the elsarticle email font}%
}
\makeatother

\usepackage{amsmath,amssymb,mathtools}
\usepackage{unicode-math}
\usepackage{textcomp}
\usepackage{siunitx}
\usepackage{graphicx}
\usepackage{booktabs,longtable,array,calc}
\usepackage{ragged2e}
\usepackage{caption}
\usepackage{float}
\usepackage[section]{placeins}

\usepackage{xcolor}
\usepackage{xurl}
\usepackage{hyperref}
\hypersetup{
  colorlinks=true,
  linkcolor=black,
  citecolor=black,
  urlcolor=blue,
  pdfcreator={XeLaTeX with elsarticle}
}

\makeatletter
\let\ps@pprintTitle\ps@plain
\makeatother

\hypersetup{
  pdftitle={A sensor-driven physics-informed neural network for real-time occupant-centric space heating control},
  pdfauthor={Yunfei Bai; Chenghao Li; Sunku Prasad Jenne; Hao Wang; Sizhe Zhang; Jihong Wang}
}

\begin{document}

\begin{frontmatter}

\title{A sensor-driven physics-informed neural network for real-time occupant-centric space heating control}

\author[warwick]{Yunfei Bai}
\author[warwick]{Chenghao Li}
\author[warwick]{Sunku Prasad Jenne}
\author[mcmaster]{Hao Wang}
\author[warwick]{Sizhe Zhang}
\author[warwick]{Jihong Wang\corref{corresponding}}
\cortext[corresponding]{Corresponding author}
\ead{jihong.wang@warwick.ac.uk}

\address[warwick]{School of Engineering, University of Warwick, Coventry CV4 7AL, United Kingdom}
\address[mcmaster]{McMaster Automotive Resource Centre, McMaster University, Hamilton, ON L8P 0A6, Canada}

\begin{abstract}
Conventional space heating systems typically rely on single or sparse fixed-point temperature measurements, overlooking non-uniform indoor temperature distributions, transient thermal dynamics, and occupant-specific comfort requirements. This study explores an occupant-centric control strategy that aims to maintain the desired temperature around the occupant location. Spatially and temporally varying temperature fields, together with occupant mobility, make real-time local temperature estimation challenging. To address this challenge, we propose a control-oriented Physics-Informed Neural Network (PINN) model for indoor temperature-field prediction. Conventional PINNs rely on explicit time inputs and assume consistent temporal patterns across samples, which limits adaptability under variable initial conditions and heating sequences. The proposed model infers temporal dynamics implicitly from spatial coordinates and limited sensor measurements, improving robustness to changes in initial conditions and external disturbances. The method was validated across nine horizontal sensing planes in three controlled room-scale configurations. Against a validated COMSOL Multiphysics simulation, the PINN predicted the evaluated temperature fields with a maximum relative error below 1.85\% across validation cases. These predictions supported an occupant-centric space heating control scheme. Compared with COMSOL-based offline control, the PINN-based system achieved an average prediction error of around 0.67\%, and its rapid inference supports real-time occupant-centric control implementation. This work represents a move away from reliance on fixed-point temperature measurements in space heating control to an occupant-centric control strategy informed by real-time environmental feedback.

\end{abstract}


\end{frontmatter}

\section{Introduction}

Heating and cooling buildings is a major global energy and carbon challenge: space and water heating represent around half of building energy use \cite{main1}. The buildings and construction sector currently accounts for 32\% of global energy use and 34\% of global \(\mathrm{CO}_{2}\) emissions \cite{main2}, while net-zero pathways require a substantial reduction in space-heating energy consumption \cite{main3}. Previous work has shown that occupant-centric heating control based on local thermal conditions can reduce energy use while maintaining comfort in a controlled room setting \cite{main4}. Building on this motivation, the present study investigates an intelligent control method for reducing residential space-heating energy use in a conditioned single-room environment while meeting occupant comfort requirements.

Space-heating control commonly relies on a single zone temperature or a small number of fixed-point measurements, which may not resolve local temperature variations or individual comfort conditions \cite{main5,main7}. Indoor temperature distributions are inherently non-uniform, and studies of spatiotemporal prediction, multi-sensor control, and sensor placement show the value of resolving these variations \cite{main8,main9,main10}. Feedback that poorly represents the occupied region can therefore lead to inefficient control and local discomfort \cite{main8,main9}. In practice, obtaining high-resolution temperature distributions within heated indoor spaces remains challenging. Dense sensor networks can improve observability but increase hardware, installation, maintenance, and data-integration burdens \cite{main7,main10}. Consequently, fast models that reconstruct indoor temperature fields from limited sensor measurements are needed for control strategies that account for both spatiotemporal variation and occupant comfort \cite{main10,main12}.

Indoor temperature fields can be obtained through either forward or inverse formulations. A forward problem predicts the temperature distribution from a physical model and specified inputs, whereas an inverse problem infers unknown quantities or reconstructs a field from observations. Computational fluid dynamics (CFD) is a well-established forward-modelling approach. Data-driven and physics-informed methods have increasingly been investigated for inverse reconstruction and surrogate modelling, with physics-informed neural networks (PINNs) offering a promising way to incorporate governing equations into learning \cite{main13}.

CFD requires specified initial and boundary conditions and offers high accuracy and physical interpretability when the governing model and system parameters are sufficiently known. However, resolving time-varying boundaries and fine spatial meshes increases the measurement, model-preparation, and computational burden \cite{main12,main15}. Consequently, high-fidelity CFD is often impractical for closed-loop real-time control applications \cite{main15}.

Purely data-driven approaches, primarily based on deep neural networks (DNNs), have gained increasing attention because of their rapid inference capabilities. By exploiting layered nonlinear mappings, DNNs can approximate complex temperature fields without solving an explicit physical model at inference. Their flexibility supports near-real-time deployment in control systems \cite{main8,main17}. However, DNNs can require substantial training data, and their accuracy depends strongly on data quality and coverage. Without explicit physical constraints, performance can degrade under operating conditions that are inadequately represented in the training data \cite{main13,main18}.

PINNs combine neural networks with governing physical laws, commonly expressed as partial differential equations (PDEs), and can address both forward and inverse problems \cite{main19,main20}. Embedding physical residuals and observations in the loss function can improve physical consistency and predictive robustness in some settings \cite{main13,main18,main22}. These advantages are not universal: optimisation becomes more difficult in high-dimensional problems, and physical constraints do not by themselves guarantee reliable extrapolation beyond the training distribution \cite{main13,main21}. Hybrid architectures have also combined convolutional feature extraction with physics-informed constraints to improve computational efficiency \cite{main16}. PINNs have been applied to fluid mechanics \cite{main23,main24,main25}, continuum and solid mechanics \cite{main14,main26}, cardiovascular flow modelling \cite{main27}, heat transfer and virtual thermal sensing \cite{main19,main22,main29}, power systems \cite{main31}, and water hydrodynamics \cite{main32}.

Within the building domain, physics-informed or physically consistent neural networks have been developed for control-oriented thermal modelling \cite{main60}, demand-response-oriented thermal modelling \cite{main18}, scalable multi-zone thermal-state modelling \cite{main62}, and fast full-field indoor temperature prediction \cite{main64}. A recent review synthesises physics-informed machine-learning approaches for building energy modelling and highlights continuing generalisation and application challenges \cite{main61}. Once trained, selected surrogate formulations can provide rapid inference while retaining physical constraints under sparse data \cite{main22,main60,main64}, motivating their use in control-oriented settings.

Space heating control is inherently dynamic because indoor temperature fields evolve with changing initial conditions, ambient disturbances, and heater-control actions. Conventional PINNs commonly include absolute time as an input, coupling the learned solution to the initial conditions and disturbance trajectories represented during training. In practical heating control, however, initial temperature fields and control sequences vary across operating episodes. A model trained on specific time histories may therefore transfer poorly to unseen temporal patterns. These considerations motivate a control-oriented formulation that infers temporal evolution from current sensor information without requiring deployment trajectories to reproduce the training histories.

Space heating controllers regulate the operating state or output of heating devices to maintain indoor thermal conditions. Thermostatic and proportional-integral-derivative (PID) controllers provide simple baseline implementations; representative implementations and broad control reviews document their use in building HVAC systems \cite{main33,main36}. More recent work has shifted towards intelligent control methods that incorporate system models, data analysis, and learning-based policies \cite{main36,main38}. Representative approaches include fuzzy logic control (FLC) \cite{main40}, model predictive control (MPC) \cite{main37}, and deep reinforcement learning (DRL) \cite{main57}. Occupant-centric control has also demonstrated combined energy and comfort benefits in a side-by-side building experiment \cite{main63}. Nevertheless, feedback in many controllers is reduced to a single zone temperature, a small set of fixed-point measurements, or a spatial average \cite{main9,main10,main57}, which can obscure the local thermal conditions experienced by a moving occupant.

In our previous studies, we developed multiphysics-based occupant-centric space heating control methods that used spatially resolved temperature predictions at the occupant's location as feedback for the heating controller \cite{main4,main58}. The evaluated scenarios showed substantial improvements in thermal comfort and indicated potential energy savings compared with conventional fixed-point temperature control. However, the underlying approach involved computationally intensive simulations and licensed commercial software, limiting its suitability for real-time deployment. These limitations motivate a sensor-informed temperature-field model that can estimate spatially varying indoor conditions in real time for occupant-centric heating control.

To address these challenges, this study develops a control-oriented PINN architecture that removes time as an explicit input. Instead, the model infers the temporal evolution of temperature fields implicitly from spatial coordinates and sparse sensor measurements. This design enhances robustness to temporal variability and reduces dependence on densely sampled temporal data. Benchmark results show that the proposed model achieves accuracy comparable to CFD while substantially reducing computation time, supporting real-time deployment. Building upon this model, this study further develops an occupant-centric space heating control framework that dynamically adjusts heating strategies based on localized temperature predictions at the occupant\textquotesingle s position, thereby improving thermal comfort and energy efficiency.

\section{Methods}

To obtain the temperature field in a heated space, CFD relies on complete physical constraints, including the laws of mass conservation, momentum conservation, and energy conservation. This allows for precise simulation of heat transfer and fluid motion characteristics within the controlled space. For transient problem formulations, the state variables that depend on both time and spatial variations primarily include the temperature field and air velocity. In experimental settings, spatial temperature data can be collected through deployed temperature sensors, as illustrated in Fig.~\ref{fig:1}(b). However, as air velocity is characterised by a three-dimensional vector field with strong nonlinearity and turbulent behaviour, achieving high-fidelity measurements of its spatial distribution in real-world scenarios remains a significant challenge for conventional sensor networks.

The advantage of PINNs lies in their ability to model temperature fields without relying entirely on complete physical constraint equations. Instead, PINNs leverage partial physical information to guide the neural network\textquotesingle s parameter updates.

\subsection{Experimental Setup, Data Acquisition, and Control-Oriented PINN Architecture}

\subsubsection{Data Acquisition for Model Training and Validation}

\begin{figure}[htbp]
\centering
\includegraphics[width=\linewidth,height=0.78\textheight,keepaspectratio]{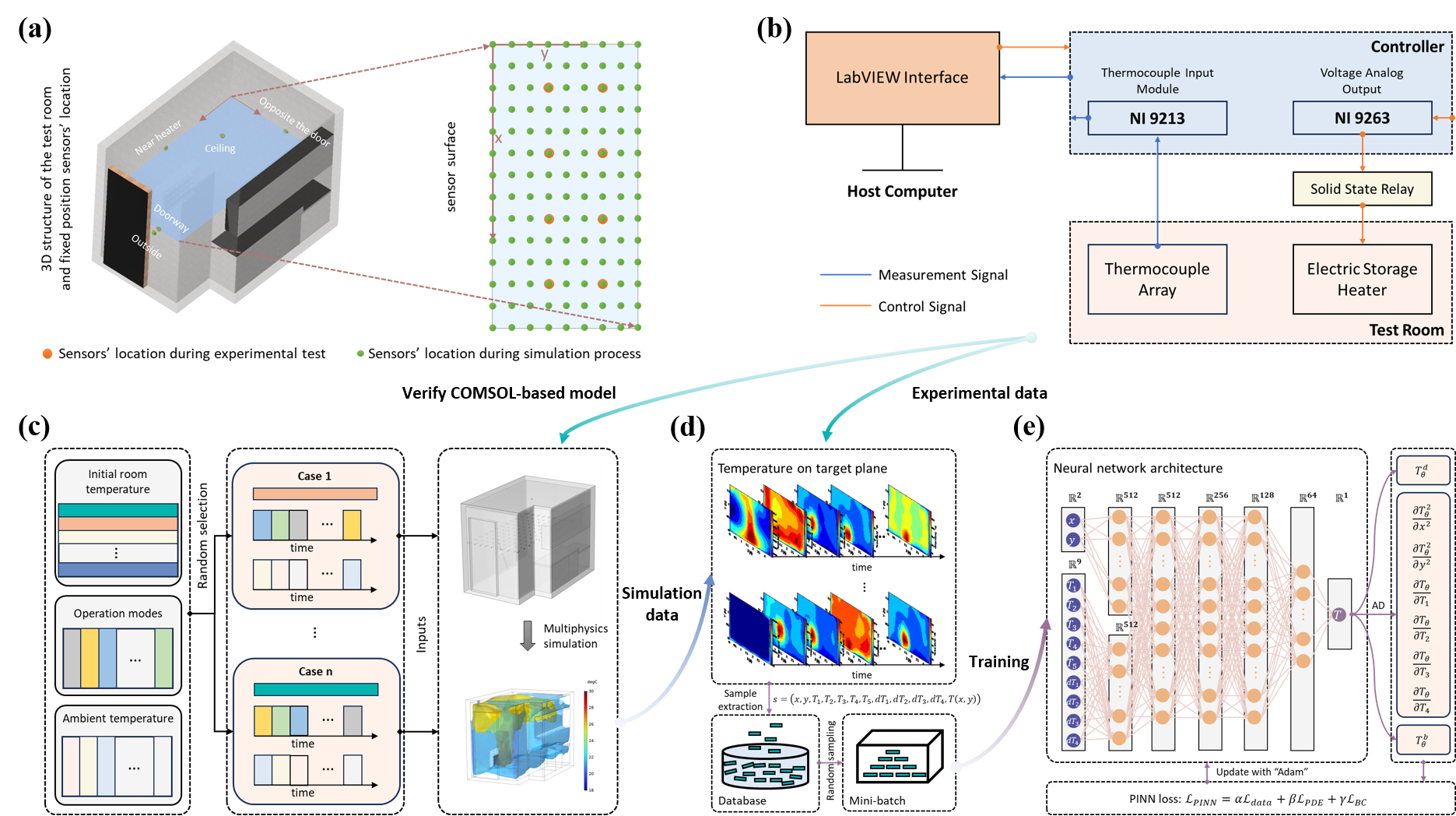}
\caption{Methodology overview of the control-oriented PINN model. (a) Locations of sensor deployment in both experimental and simulation setups. (b) Schematic of the experimental setup. (c) Simulation data generated under random initial conditions, ambient temperatures, and control modes. (d) Construction of the synthetic dataset, which consists of simulated data on the target plane and experimentally collected measurements, used for training, validation, and testing of the PINN model. (e) Architecture of the control-oriented PINN. Automatic differentiation (AD) is used to obtain the partial derivatives of the predicted temperature field with respect to temporal and spatial inputs, enabling physics-informed loss construction and facilitating parameter updates tailored for control-oriented applications.}
\label{fig:1}
\end{figure}

The dataset used for training, validation, and testing of the PINN in this study is a synthetic dataset, which integrates both experimentally collected measurements and Multiphysics simulation data, as illustrated in Fig.~\ref{fig:1}(d). The data used to train the PINN model originate from two sources: experimental measurements and Multiphysics simulation results, as shown in Fig.~\ref{fig:1}(d). The adoption of a synthetic dataset is motivated by the limitations of purely experimental data: although experiments provide accurate measurements, they are time-consuming, costly, and limited in diversity. By incorporating high-fidelity simulation data to complement the relatively sparse experimental dataset, the diversity and coverage of the dataset can be significantly enhanced.

\subsubsection{Experimental Test Environment}

A kitchen room located in the author's laboratory was selected as the test environment for this study. A storage-type electric heater with multiple operational modes was installed to provide space heating during the experiments. Ideally, a dense sensor network or high-resolution temperature sensing across the entire space would enable detailed mapping of the indoor thermal environment. However, such configurations are impractical in real-world settings due to cost and deployment constraints. Considering that occupant location is primarily represented on the horizontal plane, and to balance measurement accuracy with deployment feasibility, several horizontal planes were selected and tested independently. This design allows the spatial temperature distribution in the vicinity of occupants to be captured while reducing the complexity of sensor network deployment. One example of the selected planes is illustrated in Fig.~\ref{fig:1}(a).

Occupant thermal comfort is influenced by individual physical characteristics, including height and body shape, and by vertical temperature variation. Explicitly representing these factors for a moving occupant would substantially increase model complexity. Previous work compared the average air temperature surrounding a cylindrical human-body model with the average temperature along a corresponding vertical line and found only small differences under matched conditions \cite{main4}. Because that validation concerned a line average rather than an individual point, the horizontal-plane point temperature used here should be interpreted as a control-oriented approximation rather than a complete three-dimensional comfort metric. This approximation reduces modelling complexity but does not capture all vertical and body-resolved variations.

\subsubsection{Sensor-Based Temperature Measurements}

The experimental setup, illustrated in Fig.~\ref{fig:1}(b), is organized into three main modules: temperature signal acquisition, LabVIEW-based control interface, and control execution. A total of 12 thermocouple sensors were installed at selected locations within the room, along with one sensor positioned outside to monitor ambient conditions. All temperature measurements were transmitted to an NI CompactRIO Controller via an NI thermocouple input module. The collected temperature data is visualized in real time through a LabVIEW interface and also used for the design and execution of control strategies. Based on the sensed data, the controller generates control commands to regulate the operational mode of the storage electric heater accordingly. This test rig and its acquisition and control hardware were described and validated in our previous studies \cite{main58}; only the components directly relevant to the present data acquisition are summarised here.

The spatial distribution of temperature sensors is illustrated in Fig.~\ref{fig:1}(a). Eleven sensors were installed on the target temperature measurement plane. Eight were suspended within the interior, while three were fixed near key thermal boundaries: one adjacent to the heater, one near the doorway, and one on the wall opposite the door. The suspended array captured spatial and temporal variations across the target plane for analysis and model validation. The boundary-mounted sensors provided measurements used as PINN inputs, and their placement outside typical occupant movement paths avoided obstruction. All target-plane sensors were installed at the same height to support temperature-field reconstruction. Two supplementary sensors were installed, one on the ceiling to monitor vertical stratification and one outside the room to record ambient temperature. The layout was selected according to these measurement and deployment requirements rather than through formal optimisation.

\subsubsection{Synthetic Temperature Fields from Multiphysics Simulations}

We previously developed and experimentally validated a COMSOL Multiphysics model of the same test room \cite{main4}. In this study, the model was used to generate high-resolution temperature-field data. The previous study reports the model formulation, boundary conditions, mesh-independence analysis, and solver settings. In the present simulations, a dense array of virtual probes was placed on each target horizontal plane because these probes were not subject to the deployment constraints of physical sensors. Their layout is shown in Fig.~\ref{fig:1}(a).

The simulation model is employed to generate temperature distribution data under diverse initial conditions, ambient temperatures, and control modes. Compared to the experimental setup, the simulation model offers greater flexibility and significantly reduces time requirements. As illustrated in Fig.~\ref{fig:1}(c), for each case, once the initial state is specified, the heater control mode is randomly selected at each simulation time step. The temperature field is recorded over all time steps for each case. The simulated data are then combined with experimentally collected measurements to form a synthetic dataset, which is used for training and evaluating the PINN, improving its generalization capability.

\subsubsection{Development of the Control-Oriented PINN Model}

By approximating the time derivative term in the PDEs using temperature measurements and their rates of change obtained from sensors, the explicit dependence on time in conventional PINN architectures can be effectively eliminated. Building on this insight, this paper proposes a control-oriented PINN architecture without explicit time dependency. This modification enables the network to learn the spatiotemporal evolution of the temperature field directly from temperature values and their temporal gradients, without requiring time as an explicit input variable. The detailed network architecture is illustrated in Fig.~\ref{fig:1}(e). This architecture is a multi-input, feature-fusion neural network with two branches that separately process spatial coordinate information and measured temperature data. This setup enables specific nonlinear mappings and feature extraction based on the characteristics of each data type. Spatial information and temperature data are then fused at the feature level. Finally, several fully connected layers perform deep learning on the data features to predict spatial temperature.

\subsection{Problem Formulation and Physical Constraints}

During the conventional calculation of temperature field, the mass, momentum, and energy conservation laws are applied to the bounded physical space model to obtain the temperature distribution. For the transient problem formulation, the dependent variables of interest are temperature \(T = f(x,y,z,t)\), air velocity \(\mathbf{U} = \left\{ u(x,y,z,t),v(x,y,z,t),w(x,y,z,t) \right\}\), and pressure, where \(x\), \(y\), and \(z\) represent the coordinate position in three-dimensional space. The governing equations for the conservation of mass, momentum, and energy are described in Eqs.~\eqref{eq:6}--\eqref{eq:8}.

The continuity equation is used to describe the conservation of mass and can be written as follows:

\begin{equation}
\nabla \bullet \left( \rho\mathbf{U} \right) = 0
\label{eq:6}
\end{equation}

where, \(\rho\) is the density (\(kg/m^{3}\)), which is influenced by temperature; \(\mathbf{U}\) is the three-dimensional air velocity vector (\(m/s\)).

The Navier-Stokes momentum equation describes motion of air within the room and is represented in Eq.~\eqref{eq:7} for the three-dimensional case.

\begin{equation}
\frac{\partial\left( \rho\mathbf{U} \right)}{\partial t} + \rho\left( \mathbf{U} \bullet \nabla \right)\mathbf{U =}\nabla \bullet \left\lbrack - p\mathbf{I} + \mu(\nabla\mathbf{U +}\left( \nabla\mathbf{U} \right)'\mathbf{)} \right\rbrack + \rho\mathbf{g}
\label{eq:7}
\end{equation}

where, \(p\) is air pressure (\(Pa\)). \(\mathbf{I}\) is identity matrix; \(\mu\) is the air dynamic viscosity (\(Pa \bullet s\)); \(\mathbf{g}\) is the gravity vector (\(m/s^{2}\)).

In heat transfer analysis, the energy balance equation can be expressed in the following form:

\begin{equation}
\frac{\partial\left( \rho C_{p}T \right)}{\partial t} + \rho C_{p}\mathbf{U} \bullet \nabla T = \nabla \bullet (k\nabla T) + Q
\label{eq:8}
\end{equation}

where, \(C_{p}\) is the specific heat at constant pressure (\(J/(kg \cdot K)\)), which is influenced by temperature; \(k\) is the thermal conductivity (\(W/(m \cdot K)\)); \(Q\) is the applied external heat source (\(W/m^{3}\)).

As discussed previously, air velocity is a three-dimensional variable and is difficult to measure accurately in practical scenarios. Therefore, by neglecting the effect of air velocity, the governing equations can be simplified. For the heat transfer problem on the plane shown in Fig.~\ref{fig:1}(a), the internal space is defined as the domain \(\Omega\), with its boundary denoted by \(\partial\Omega\). The partial differential equations (Eqs.~\eqref{eq:6}--\eqref{eq:8}) governing the heat transfer within \(\Omega\) can therefore be simplified and expressed as:

\begin{equation}
\rho C_{p}\frac{\partial T}{\partial t} - k\left( \frac{\partial^{2}T}{\partial x^{2}} + \frac{\partial^{2}T}{\partial y^{2}} \right) = 0
\label{eq:9}
\end{equation}

where, \(\rho\) is the density (\(kg/m^{3}\)); \(C_{p}\) is the specific heat capacity (\(J/(kg \cdot \si{\celsius})\)); \(T\) is the temperature (\(\si{\celsius}\)); \(t\) is the time (\(s\)); \(k\) is the thermal conductivity (\(W/(m \cdot \si{\celsius})\)); \(x\), \(y\) represent the coordinate position on this plane (\(m)\).

To address such unsteady heat transfer problems, the inputs of the PINN are typically defined as spatial coordinates \((x,y)\) and discrete time points \(t\), while the output corresponds to the temperature \(T\) at the specified coordinates and time. During the training process, automatic differentiation (AD) is employed to compute the second-order derivatives of \(T\) with respect to \(x\) and \(y\), as well as the first-order time derivative of \(t\). These derivatives are then substituted into Eq.~\eqref{eq:9} to evaluate the residual, which is incorporated into the PINN loss function, guiding the network parameter updates in a physically consistent manner.

Such a PINN is applied in scenarios where the initial conditions of the modeling object and the control time sequence remain consistent across the training and application phases. However, in practical control systems, the initial conditions of the modelled object can vary at the start of each control cycle. Control actions are dynamically determined based on feedback from sensor measurements, resulting in control sequences that are not fixed over time. To develop a PINN model that can be integrated into a control system, discrete time \(t\) cannot be used as an input to the network, and the time derivative of the output temperature \(T\) needs to be approximated by alternative means.

In the heated room, a small number of temperature sensors can be strategically deployed without interfering with normal occupant activities. In this study, four temperature sensors are installed within the room, as illustrated in Fig.~\ref{fig:1}(a). These sensors continuously record the temporal evolution of temperature. As the temperature varies only with time at each sensor location, the time derivative is approximated using a first-order finite difference method, as shown in Eq.~\eqref{eq:10}.

\begin{equation}
\frac{dT_{S_{n}}}{dt} \approx \frac{T_{S_{n}}(t) - T_{S_{n}}(t - \Delta{}t)}{\Delta{}t},\ \ \ n\epsilon\left\{ 1,2,3,4 \right\}
\label{eq:10}
\end{equation}

where, \(T_{S_{n}}\) represents the temperature measured by the \(n\)th sensor (\(\si{\celsius}\)), and \(\Delta{}t\) denotes the time interval between two consecutive measurements (\(s\)).

Since the predicted temperature depends on the time-varying sensor inputs, the time derivative can be approximated using the chain rule. Specifically, AD is used to compute the sensitivities of the predicted temperature with respect to each sensor input, which are then combined with the temporal rates of sensor readings, as shown in Eq.~\eqref{eq:11}.

\begin{equation}
\frac{\partial T}{\partial t} \approx \sum_{n = 1}^{4}{\frac{\partial T}{\partial T_{S_{n}}} \bullet}\frac{dT_{S_{n}}}{dt}
\label{eq:11}
\end{equation}

During the training process of the PINN, the PDE error is defined as the residual of Eq.~\eqref{eq:9}. The approximate time derivative calculated in Eq.~\eqref{eq:11} is incorporated into the PDE error to account for the time derivative component. In this way, even without explicit time information, the model can approximate the time derivative by leveraging the sensor temperature data and its rate of change, allowing the model to learn a realistic evolution of the temperature field. By combining Eqs.~\eqref{eq:9} and \eqref{eq:11}, the PDE error can be calculated as follows:

\begin{equation}
\mathcal{L}_{PDE} = \rho C_{p}\sum_{n = 1}^{4}{\frac{\partial T}{\partial T_{S_{n}}} \bullet \frac{dT_{S_{n}}}{dt} - k\left( \frac{\partial^{2}T}{\partial x^{2}} + \frac{\partial^{2}T}{\partial y^{2}} \right)}
\label{eq:12}
\end{equation}

The proposed control-oriented PINN architecture is shown in Fig.~\ref{fig:1}(e). In this architecture, the temperature \(T\) at any location \((x,y)\) can be predicted using the PINN. The predicted temperature is expressed in Eq.~\eqref{eq:13}.

\begin{equation}
T \approx T_{PINN}\left( \mathbf{X};\theta \right)
\label{eq:13}
\end{equation}

where, \(T_{PINN}\) represents the temperature predicted by the PINN. \(\mathbf{X}\) is a vector composed of spatial coordinates and observed temperatures. \(\theta\) denotes the set of PINN parameters, including weights and biases.

The hyperparameters of the proposed PINN network are detailed in Table~\ref{tab:3}. These parameters were selected based on a trial-and-error process. Although they may not be globally optimal, they enable the network to effectively learn the underlying complex patterns and achieve satisfactory performance. The learning rate is set to decay exponentially, as described in Eq.~\eqref{eq:14}.

\begin{equation}
\delta_{i} = \delta_{0}\gamma^{\frac{i}{s}} = \delta_{0}e^{-\lambda i},
\qquad \lambda = -\frac{\ln\gamma}{s}
\label{eq:14}
\end{equation}

where, \(\delta_{i}\) represents the learning rate at iteration \(i\); \(\delta_{0}\) is the initial learning rate; \(\gamma\), with \(0<\gamma<1\), is the decay rate; \(s\) is the decay step; and \(\lambda=-\ln(\gamma)/s>0\) is the equivalent per-iteration exponential decay constant.

\begingroup
\small
\setlength{\tabcolsep}{4pt}
\begin{longtable}[]{@{}
  >{\raggedright\arraybackslash}p{(\linewidth - 4\tabcolsep) * \real{0.1410}}
  >{\raggedright\arraybackslash}p{(\linewidth - 4\tabcolsep) * \real{0.5975}}
  >{\raggedright\arraybackslash}p{(\linewidth - 4\tabcolsep) * \real{0.2615}}@{}}
\caption{Hyperparameters of the PINN.}\label{tab:3}\tabularnewline
\toprule\noalign{}
\begin{minipage}[b]{\linewidth}\raggedright
\textbf{No.}
\end{minipage} & \begin{minipage}[b]{\linewidth}\raggedright
\textbf{Hyperparameters}
\end{minipage} & \begin{minipage}[b]{\linewidth}\raggedright
\textbf{Values}
\end{minipage} \\
\midrule\noalign{}
\endfirsthead
\toprule\noalign{}
\begin{minipage}[b]{\linewidth}\raggedright
\textbf{No.}
\end{minipage} & \begin{minipage}[b]{\linewidth}\raggedright
\textbf{Hyperparameters}
\end{minipage} & \begin{minipage}[b]{\linewidth}\raggedright
\textbf{Values}
\end{minipage} \\
\midrule\noalign{}
\endhead
\bottomrule\noalign{}
\endlastfoot
\begin{minipage}[t]{\linewidth}\raggedright
1
\end{minipage} & \begin{minipage}[t]{\linewidth}\raggedright
Number of neurons in input layer of Branch 1
\end{minipage} & \begin{minipage}[t]{\linewidth}\raggedright
2
\end{minipage} \\
\begin{minipage}[t]{\linewidth}\raggedright
2
\end{minipage} & \begin{minipage}[t]{\linewidth}\raggedright
Number of neurons in input layer of Branch 2
\end{minipage} & \begin{minipage}[t]{\linewidth}\raggedright
9
\end{minipage} \\
\begin{minipage}[t]{\linewidth}\raggedright
3
\end{minipage} & \begin{minipage}[t]{\linewidth}\raggedright
Number of neurons in the output layer
\end{minipage} & \begin{minipage}[t]{\linewidth}\raggedright
1
\end{minipage} \\
\begin{minipage}[t]{\linewidth}\raggedright
4
\end{minipage} & \begin{minipage}[t]{\linewidth}\raggedright
Number of neurons in fully connected layers of Branch 1
\end{minipage} & \begin{minipage}[t]{\linewidth}\raggedright
512
\end{minipage} \\
\begin{minipage}[t]{\linewidth}\raggedright
5
\end{minipage} & \begin{minipage}[t]{\linewidth}\raggedright
Number of neurons in fully connected layers of Branch 2
\end{minipage} & \begin{minipage}[t]{\linewidth}\raggedright
512
\end{minipage} \\
\begin{minipage}[t]{\linewidth}\raggedright
6
\end{minipage} & \begin{minipage}[t]{\linewidth}\raggedright
Size of the hidden layer after fusion
\end{minipage} & \begin{minipage}[t]{\linewidth}\raggedright
4
\end{minipage} \\
\begin{minipage}[t]{\linewidth}\raggedright
7
\end{minipage} & \begin{minipage}[t]{\linewidth}\raggedright
Number of neurons in the first layer after fusion
\end{minipage} & \begin{minipage}[t]{\linewidth}\raggedright
512
\end{minipage} \\
\begin{minipage}[t]{\linewidth}\raggedright
8
\end{minipage} & \begin{minipage}[t]{\linewidth}\raggedright
Number of neurons in the second layer after fusion
\end{minipage} & \begin{minipage}[t]{\linewidth}\raggedright
256
\end{minipage} \\
\begin{minipage}[t]{\linewidth}\raggedright
9
\end{minipage} & \begin{minipage}[t]{\linewidth}\raggedright
Number of neurons in the third layer after fusion
\end{minipage} & \begin{minipage}[t]{\linewidth}\raggedright
128
\end{minipage} \\
\begin{minipage}[t]{\linewidth}\raggedright
10
\end{minipage} & \begin{minipage}[t]{\linewidth}\raggedright
Number of neurons in the fourth layer after fusion
\end{minipage} & \begin{minipage}[t]{\linewidth}\raggedright
64
\end{minipage} \\
\begin{minipage}[t]{\linewidth}\raggedright
11
\end{minipage} & \begin{minipage}[t]{\linewidth}\raggedright
Initial learning rate
\end{minipage} & \begin{minipage}[t]{\linewidth}\raggedright
0.001
\end{minipage} \\
\begin{minipage}[t]{\linewidth}\raggedright
12
\end{minipage} & \begin{minipage}[t]{\linewidth}\raggedright
Learning rate decay function
\end{minipage} & \begin{minipage}[t]{\linewidth}\raggedright
Exponential decay
\end{minipage} \\
\begin{minipage}[t]{\linewidth}\raggedright
13
\end{minipage} & \begin{minipage}[t]{\linewidth}\raggedright
Decay rate
\end{minipage} & \begin{minipage}[t]{\linewidth}\raggedright
0.1
\end{minipage} \\
\begin{minipage}[t]{\linewidth}\raggedright
14
\end{minipage} & \begin{minipage}[t]{\linewidth}\raggedright
Decay steps
\end{minipage} & \begin{minipage}[t]{\linewidth}\raggedright
5000 iterations
\end{minipage} \\
\begin{minipage}[t]{\linewidth}\raggedright
15
\end{minipage} & \begin{minipage}[t]{\linewidth}\raggedright
Optimizer
\end{minipage} & \begin{minipage}[t]{\linewidth}\raggedright
Adam
\end{minipage} \\
\begin{minipage}[t]{\linewidth}\raggedright
16
\end{minipage} & \begin{minipage}[t]{\linewidth}\raggedright
Activation function
\end{minipage} & \begin{minipage}[t]{\linewidth}\raggedright
Tanh
\end{minipage} \\
\end{longtable}
\endgroup

\subsection{Physics-Informed Loss Function Design}

The training data set for the proposed PINN model consists of multiple experimental and simulation cases, each with distinct initial conditions. Moreover, since the proposed PINN architecture does not rely on explicit time as an input parameter, the training data does not include time labels. Under this condition, the traditional initial condition constraints used in conventional PINNs, which are typically time-dependent, cannot be directly incorporated into the loss function. Therefore, the loss function of the proposed PINN is composed of three primary components: PDE residual loss, data fidelity loss, and boundary condition loss. These components ensure that the learned temperature field adheres to the governing physical equations, matches the measured sensor data, and satisfies the boundary constraints, respectively. The loss function is expressed in Eq.~\eqref{eq:15}.

\begin{equation}
\mathcal{L}\left( \mathbf{X};\theta \right) = \alpha_{PDE}\mathcal{L}_{PDE}\left( \mathbf{X};\theta \right) + \alpha_{data}\mathcal{L}_{data}\left( \mathbf{X};\theta \right) + \alpha_{BC}\mathcal{L}_{BC}\left( \mathbf{X};\theta \right)
\label{eq:15}
\end{equation}

where, \(\mathcal{L}\left( \mathbf{X};\theta \right)\) represents the total loss of the PINN. \(\mathcal{L}_{PDE}\left( \mathbf{X};\theta \right)\), \(\mathcal{L}_{data}\left( \mathbf{X};\theta \right)\) and \(\mathcal{L}_{BC}\left( \mathbf{X};\theta \right)\) are the PDE loss, data error in \(\Omega\), and data error on \(\partial\Omega\), respectively. \(\alpha_{PDE}\), \(\alpha_{data}\) and \(\alpha_{BC}\) denote the weighing coefficients for each corresponding loss or error term.

The error between the predicted temperature data inside the targeted plane and the actual data is defined in Eq.~\eqref{eq:16}.

\begin{equation}
\mathcal{L}_{data}\left( \mathbf{X};\theta \right) = \frac{1}{N}\sum_{i = 1}^{N}{(T_{PINN}\left( \mathbf{X}_{\mathbf{i}};\theta \right) - T_{data}(x_{i},y_{i}))}^{2},\ \left( x_{i},y_{i} \right) \in \Omega
\label{eq:16}
\end{equation}

where, \(N\) represents the number of samples in the mini batch belonging to \(\Omega\). \(T_{data}(x_{i},y_{i})\) denotes the actual temperature at the location \(\left( x_{i},y_{i} \right)\).

For the physics loss term, the heat conduction PDE is used to calculate the physical consistency of the predictions. Based on Eq.~\eqref{eq:12}, this part is defined in Eq.~\eqref{eq:17}.

\begin{equation}
\begin{aligned}
\mathcal{L}_{PDE}\left(\mathbf{X};\theta\right)
&= \frac{1}{M}\sum_{j=1}^{M}\Biggl[
\rho C_{p}\sum_{n=1}^{4}
\frac{\partial T_{PINN}\left(\mathbf{X}_{\mathbf{j}};\theta\right)}{\partial T_{S_{n}}}
\bullet \frac{dT_{S_{n}}}{dt} \\
&\quad {}- k\left(
\frac{\partial^{2}T_{PINN}\left(\mathbf{X}_{\mathbf{j}};\theta\right)}{\partial x_{j}^{2}}
+ \frac{\partial^{2}T_{PINN}\left(\mathbf{X}_{\mathbf{j}};\theta\right)}{\partial y_{j}^{2}}
\right)\Biggr]^{2},
\quad \left(x_{j},y_{j}\right) \in \Omega
\end{aligned}
\label{eq:17}
\end{equation}

where, \(M\) represents the number of samples in the mini batch that are required to satisfy the PDE equation.

Due to the uncertainty and potential noise in the boundary condition measurements, this study defines the boundary loss as the error between the PINN-predicted data and the collected boundary data, as shown in Eq.~\eqref{eq:18}.

\begin{equation}
\mathcal{L}_{BC}\left( \mathbf{X};\theta \right) = \frac{1}{V}\sum_{k = 1}^{V}{{(T_{PINN}\left( \mathbf{X}_{\mathbf{k}};\theta \right) - T_{data}(x_{k},y_{k}))}^{2},}\ \left( x_{k},y_{k} \right) \in \partial\Omega
\label{eq:18}
\end{equation}

where, \(V\) represents the number of samples in the mini batch belonging to \(\partial\Omega\). \(T_{data}(x_{k},y_{k})\) denotes the actual temperature at the location \(\left( x_{k},y_{k} \right)\).

During network training, gradients of the loss function are obtained by backpropagation, and the network parameters are optimised using Adam \cite{main59}. Equation~\eqref{eq:19} presents the generic gradient-based update, while the implementation uses Adam's adaptive moment estimates.

\begin{equation}
\theta_{t + 1} = \theta_{t} - \delta \bullet \nabla\mathcal{\ L}\left( \mathbf{X};\theta_{t} \right)
\label{eq:19}
\end{equation}

where, \(\theta_{t}\) is the network parameter at iteration \(t\). \(\delta\) is the learning rate calculated by Eq.~\eqref{eq:14}. \(\nabla\mathcal{\ L}\left( \mathbf{X};\theta_{t} \right)\) is the gradient of the loss function with respect to the parameters at iteration \(t\). \(\theta_{t + 1}\) is the updated network parameter for the next iteration.

Training was terminated when either the predefined maximum number of iterations was reached or the error on the internal monitoring subset fell below the specified threshold. The resulting network was then used in the control system.

\subsection{Training and Evaluation Design}

The evaluation was organised in three stages. First, the COMSOL Multiphysics model was compared with experimental measurements. The validated simulation model and experimental measurements were then used to construct the data for PINN optimisation. Finally, the trained PINN was evaluated on separately generated temperature-field cases and within a matched occupant-centric heating-control scenario.

\begin{figure}[H]
\centering
\includegraphics[width=\linewidth,height=0.78\textheight,keepaspectratio]{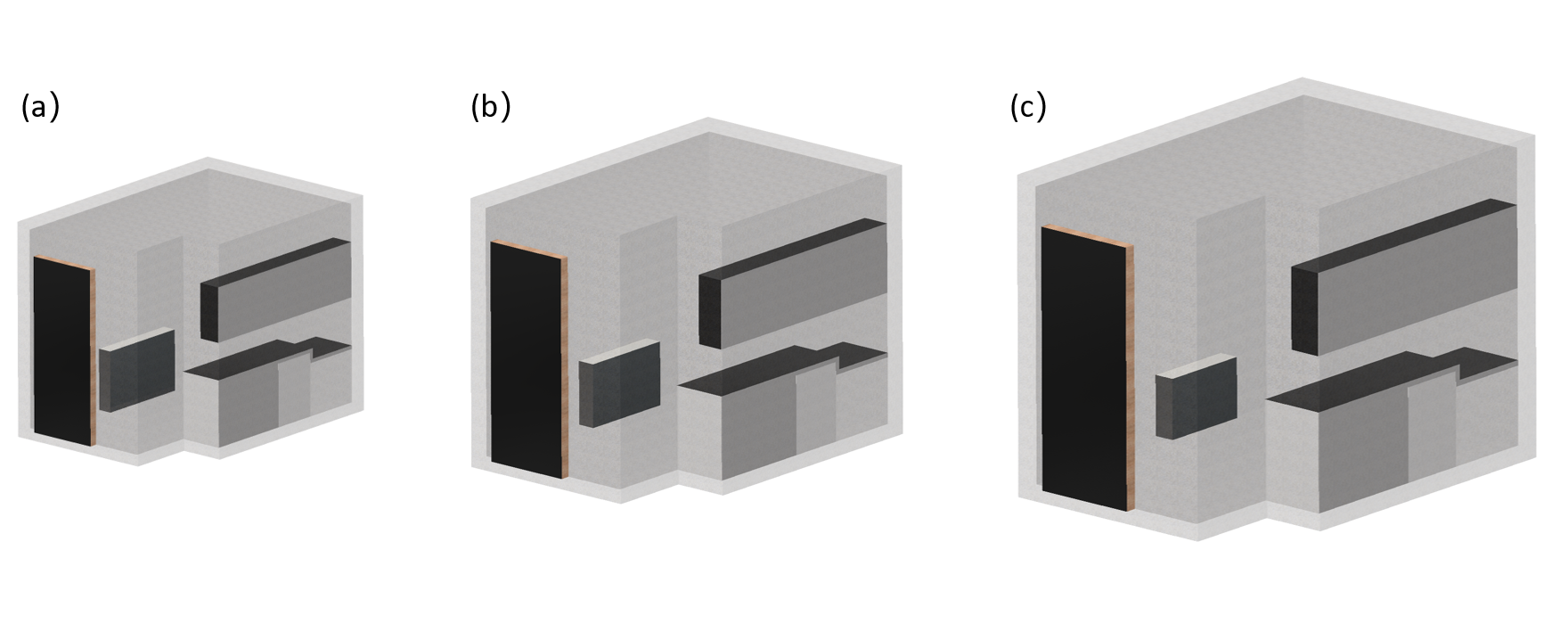}
\caption{Room configurations. (a) 80\% scaled; (b) baseline; (c) 120\% scaled.}
\label{fig:2}
\end{figure}

To examine robustness under controlled room-scale geometric variation, three room configurations were investigated, as illustrated in Fig.~\ref{fig:2}. Fig.~\ref{fig:2}(b) is based on a real room, while Fig.~\ref{fig:2}(a) and Fig.~\ref{fig:2}(c) are obtained by scaling the dimensions of Fig.~\ref{fig:2}(b) (excluding the heater) to 80\% and 120\%, respectively. Although the three rooms share similar structural characteristics, they provide a controlled setup for evaluating the sensitivity of the proposed PINN model to changes in spatial scale.

\begin{figure}[H]
\centering
\includegraphics[width=\linewidth,height=0.78\textheight,keepaspectratio]{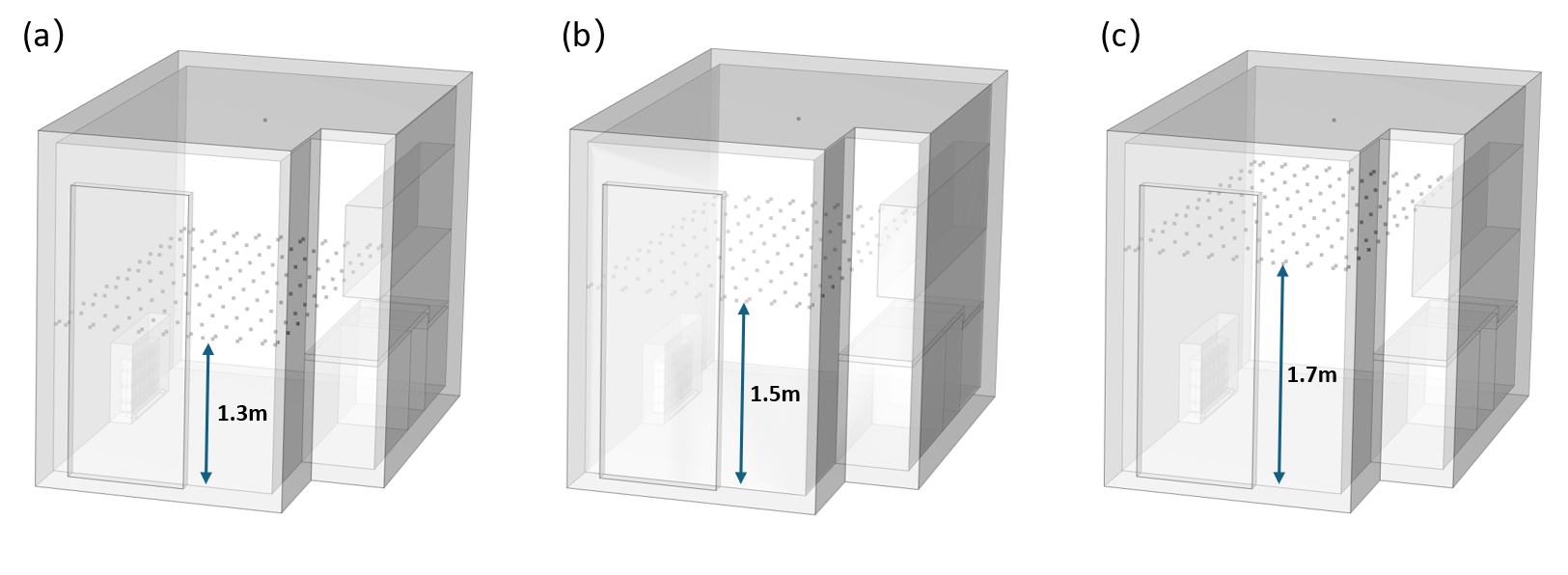}
\caption{Horizontal planes at different heights. (a) 1.3 m above the floor; (b) 1.5 m above the floor; (c) 1.7 m above the floor.}
\label{fig:3}
\end{figure}

For the room shown in Fig.~\ref{fig:2}(b), three horizontal planes located at heights of 1.3 m, 1.5 m, and 1.7 m above the floor were selected, as illustrated in Fig.~\ref{fig:3}. For each plane, both experimental measurements and simulation data were collected, and a combined synthetic dataset was used to train and validate the PINN model.

For the scaled rooms in Fig.~\ref{fig:2}(a) and Fig.~\ref{fig:2}(c), corresponding horizontal planes were defined according to the scaling ratios. For each plane, simulation data were generated and used for training and validation of the PINN model. In total, the proposed PINN framework was evaluated across nine planes in three room configurations.

These configurations were selected to test whether the control-oriented PINN remains accurate when the sensing-plane height and overall room scale change in a controlled manner. They should therefore be interpreted as room-scale robustness tests within the evaluated geometry family.

\subsubsection{Multiphysics Validation Protocol and Error Metrics}

Based on the established experimental testing platform, this study conducted tests for six heater operation scenarios for each horizontal plane, including three single-mode tests and three mode-switching tests. The single-mode tests involved the heater operating at 1 kW, 2 kW, and 3 kW, respectively. The mode-switching tests involved the heater operating at 3 kW for a period before switching to 2 kW, 3 kW before switching to 1 kW, and 2 kW before switching to 1 kW. Each test lasted for one hour.

The Multiphysics simulation model was configured with initial room conditions and heater time-control sequences corresponding to the six test scenarios to simulate the evolution of temperature distribution in the room under these settings. Simulated temperature data were extracted at 12 locations. These locations included the positions of four fixed indoor temperature sensors during the experimental tests, and eight additional suspended temperature sensor positions in the space, as illustrated in Fig.~\ref{fig:1}(a).

The relative and absolute errors between the experimental and simulated temperatures are defined in Eqs.~\eqref{eq:1} and \eqref{eq:2}, respectively.

\begin{equation}
\Psi_{Nerr}\left. \ (t) \right|_{sp(k)} = \frac{\left| \left. \ T(t) \right|_{sp(k)}^{experiment} - \left. \ T(t) \right|_{sp(k)}^{simulation} \right|}{\left. \ T(t) \right|_{sp(k)}^{experiment}} \times 100\%
\label{eq:1}
\end{equation}

\begin{equation}
\phi_{Nerr}\left. \ (t) \right|_{sp(k)} = \left| \left. \ T(t) \right|_{sp(k)}^{experiment} - \left. \ T(t) \right|_{sp(k)}^{simulation} \right|
\label{eq:2}
\end{equation}

where, \(k\) is the position of the sensor, \(k \in \left\{ 1,2,3,\ldots,12 \right\}\); \(\Psi_{Nerr}\left. \ (t) \right|_{sp(k)}\) is the relative accuracy at the \(k\)th sensor position at time \(t\); \(\phi_{Nerr}\left. \ (t) \right|_{sp(k)}\) is the absolute accuracy at the \(k\)th sensor position at time \(t\); \(\left. \ T(t) \right|_{sp(k)}^{experiment}\) is the experimental data (\(\si{\celsius}\)) at the \(k\)th sensor position at time \(t\); \(\left. \ T(t) \right|_{sp(k)}^{simulation}\) is the simulation data (\(\si{\celsius}\)) at the \(k\)th sensor position at time \(t.\)

\subsubsection{PINN Training Protocol}

The data used for model optimisation comprised multiple simulation and experimental cases. Within each case, 90\% of samples formed the optimisation subset and 10\% formed an internal monitoring subset. Mini-batches were randomly drawn from the optimisation subsets to update the network parameters. Because neighbouring time steps within a case are temporally correlated, the monitoring subset was not treated as an independent test of generalisation. Independent performance evaluation used the ten separately generated dynamic cases described in the following subsection.

In Eq.~\eqref{eq:15}, the loss function of the PINN comprises three components: the PDE loss weight \(\alpha_{PDE}\), the data loss weight \(\alpha_{data}\), and the boundary loss weight \(\alpha_{BC}\), which were set to 0.05, 0.8, and 0.2, respectively. These values were determined empirically through multiple trials, with the aim of prioritising data fitting while maintaining a balanced contribution from the physical and boundary constraints. Preliminary experiments indicated that this combination provides stable training behaviour and satisfactory predictive performance.

\subsubsection{PINN Validation Protocol and Error Metrics}

Ten independently generated cases were used to evaluate the trained PINN in predicting temperature fields. These cases represented different dynamic conditions, including varying ambient temperature profiles and randomly selected heater operating modes. Detailed heater-power and ambient-temperature profiles for the ten cases are provided in the Supplementary Materials. Each case was generated using the Multiphysics simulation model and covered 3 hours at a 1-minute interval, resulting in 180 time steps. The trained PINN predicted the temperature field on the target plane at each step. The corresponding average relative and absolute errors were calculated using Eqs.~\eqref{eq:3} and \eqref{eq:4}.

\begin{equation}
\chi_{PINN}(t) = \frac{1}{M}\sum_{i = 1}^{M}{\frac{\left| \left. \ T(t) \right|_{sp(i)}^{PINN} - \left. \ T(t) \right|_{sp(i)}^{comsol} \right|}{\left. \ T(t) \right|_{sp(i)}^{comsol}} \times 100\%}
\label{eq:3}
\end{equation}

\begin{equation}
\tau_{PINN}(t) = \frac{1}{M}\sum_{i = 1}^{M}\left| \left. \ T(t) \right|_{sp(i)}^{PINN} - \left. \ T(t) \right|_{sp(i)}^{comsol} \right|
\label{eq:4}
\end{equation}

where, \(i\) is the position of probe shown in Fig.~\ref{fig:1}(a), \(i \in \left\{ 1,2,3,\ldots,M \right\}\); \(M\) is the number of probes; \(\chi_{PINN}(t)\) is the average relative accuracy at time \(t\); \(\tau_{PINN}(t)\) is the average absolute accuracy at time \(t\); \(\left. \ T(t) \right|_{sp(k)}^{PINN}\) is the predicted temperature data (\(\si{\celsius}\)) at the \(i\)th probe position at time \(t\); \(\left. \ T(t) \right|_{sp(k)}^{comsol}\) is the COMSOL simulation data (\(\si{\celsius}\)) at the \(i\)th sensor position at time \(t.\)

\subsubsection{Occupant-Centric Control Evaluation}

To achieve the shift from traditional ``space-averaged control'' to ``occupant-centric thermal comfort control'', it is crucial to leverage real-time occupant location information to accurately capture temperature variations at the occupant\textquotesingle s position, thereby optimizing heating strategies. The control-oriented PINN architecture proposed in this study enables efficient prediction of real-time temperature dynamics at the occupant\textquotesingle s location by utilizing spatial coordinates and limited temperature sensor feedback. In the model validation analysis, capturing fine-grained spatial details of the temperature field requires deploying 108 temperature probes during the simulation on selected planes within the space. In contrast, the proposed PINN-based reconstruction approach requires only four temperature sensors located at the room boundaries. This leads to a substantial reduction in sensing requirements, while the boundary-based deployment minimises intrusion and ensures compatibility with occupants\textquotesingle{} daily activities.

In the control framework, the hierarchical state machine operates as a bang-bang controller, with the objective of maintaining the temperature within the range of 22 °C to 23 °C around the occupant's location. Although bang-bang control is not considered an advanced control strategy, it is widely adopted in practical applications due to its simplicity and robustness. Therefore, it is employed in this study to evaluate the performance of the proposed PINN model within a realistic control framework. A 3-hour simulation test was conducted using both the proposed PINN-based space heating control system and the previously developed COMSOL-based space heating control system, based on a randomly selected occupant activity trajectory.

The relative numerical error between the occupant-surrounding temperatures predicted by the PINN and COMSOL during the control process was quantified using Eq.~\eqref{eq:5}.

\begin{equation}
\Psi_{control}(t) = \frac{\left| \left. \ T(t) \right|_{PINN} - \left. \ T(t) \right|_{comsol} \right|}{\left. \ T(t) \right|_{comsol}} \times 100\%
\label{eq:5}
\end{equation}

where, \(\Psi_{control}(t)\) is the relative numerical error at time \(t\); \(\left. \ T(t) \right|_{PINN}\) is the PINN predicted temperature data (\(\si{\celsius}\)) at time \(t\); \(\left. \ T(t) \right|_{comsol}\) is the COMSOL simulation data (\(\si{\celsius}\)) at time \(t.\)

This study utilised three software platforms: COMSOL Multiphysics 6.2, MATLAB R2025b, and Autodesk Inventor Professional 2025. All simulations were performed on computers equipped with AMD Ryzen 5 5600H processors (3.30 GHz), 16 GB of RAM, an NVIDIA GeForce GTX 1650 GPU, and a 64-bit Windows operating system.

\section{Results and Discussion}

\subsection{Multiphysics Simulation Model Validation}

\begin{figure}[htbp]
\centering
\includegraphics[width=\linewidth,height=0.78\textheight,keepaspectratio]{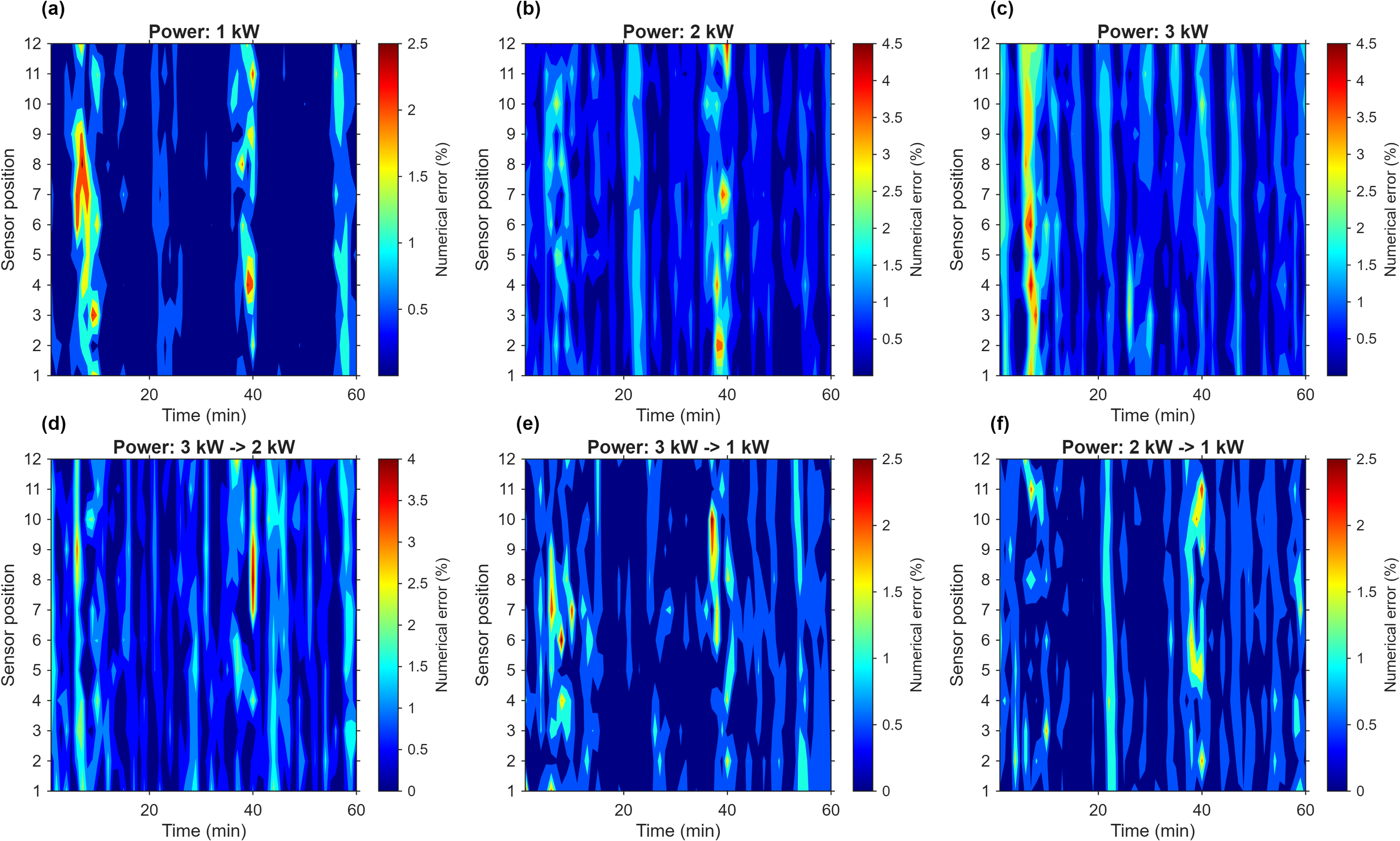}
\caption{Experimental validation of the Multiphysics simulation model. (a) 1 kW; (b) 2 kW; (c) 3 kW; (d) 3 kW $\rightarrow$ 2 kW; (e) 3 kW $\rightarrow$ 1 kW; (f) 2 kW $\rightarrow$ 1 kW. Positions 1-4 are fixed position sensors, and positions 5-12 are spatial position sensors as shown in Fig.~\ref{fig:1}(a).}
\label{fig:4}
\end{figure}

Experimental and simulated temperatures on the plane 1.5 m above the floor were compared across the six heater-operation scenarios using the metrics defined in the Methods. The time-varying errors at all 12 sensor positions are shown in Fig.~\ref{fig:4}, and the maximum and average relative and absolute errors are summarised in Table~\ref{tab:1}. The maximum and average relative errors remained below 4.60\% and 0.95\%, respectively. Corresponding results for the planes at 1.3 m and 1.7 m are provided in the Supplementary Materials.

\begin{longtable}[]{@{}
  >{\raggedright\arraybackslash}p{(\linewidth - 12\tabcolsep) * \real{0.3226}}
  >{\raggedright\arraybackslash}p{(\linewidth - 12\tabcolsep) * \real{0.1032}}
  >{\raggedright\arraybackslash}p{(\linewidth - 12\tabcolsep) * \real{0.1032}}
  >{\raggedright\arraybackslash}p{(\linewidth - 12\tabcolsep) * \real{0.1032}}
  >{\raggedright\arraybackslash}p{(\linewidth - 12\tabcolsep) * \real{0.1226}}
  >{\raggedright\arraybackslash}p{(\linewidth - 12\tabcolsep) * \real{0.1226}}
  >{\raggedright\arraybackslash}p{(\linewidth - 12\tabcolsep) * \real{0.1226}}@{}}
\caption{Experimental validation error values across different cases.}\label{tab:1}\tabularnewline
\toprule\noalign{}
\begin{minipage}[b]{\linewidth}\raggedleft
Cases

Error
\end{minipage} & \begin{minipage}[b]{\linewidth}\raggedright
1kW
\end{minipage} & \begin{minipage}[b]{\linewidth}\raggedright
2kW
\end{minipage} & \begin{minipage}[b]{\linewidth}\raggedright
3kW
\end{minipage} & \begin{minipage}[b]{\linewidth}\raggedright
3kW$\rightarrow$ 2kW
\end{minipage} & \begin{minipage}[b]{\linewidth}\raggedright
3kW$\rightarrow$ 1kW
\end{minipage} & \begin{minipage}[b]{\linewidth}\raggedright
2kW$\rightarrow$ 1kW
\end{minipage} \\
\midrule\noalign{}
\endfirsthead
\toprule\noalign{}
\begin{minipage}[b]{\linewidth}\raggedleft
Cases

Error
\end{minipage} & \begin{minipage}[b]{\linewidth}\raggedright
1kW
\end{minipage} & \begin{minipage}[b]{\linewidth}\raggedright
2kW
\end{minipage} & \begin{minipage}[b]{\linewidth}\raggedright
3kW
\end{minipage} & \begin{minipage}[b]{\linewidth}\raggedright
3kW$\rightarrow$ 2kW
\end{minipage} & \begin{minipage}[b]{\linewidth}\raggedright
3kW$\rightarrow$ 1kW
\end{minipage} & \begin{minipage}[b]{\linewidth}\raggedright
2kW$\rightarrow$ 1kW
\end{minipage} \\
\midrule\noalign{}
\endhead
\bottomrule\noalign{}
\endlastfoot
\begin{minipage}[t]{\linewidth}\raggedright
Maximum relative error (\%)
\end{minipage} & \begin{minipage}[t]{\linewidth}\raggedright
2.69
\end{minipage} & \begin{minipage}[t]{\linewidth}\raggedright
4.60
\end{minipage} & \begin{minipage}[t]{\linewidth}\raggedright
4.55
\end{minipage} & \begin{minipage}[t]{\linewidth}\raggedright
4.24
\end{minipage} & \begin{minipage}[t]{\linewidth}\raggedright
2.87
\end{minipage} & \begin{minipage}[t]{\linewidth}\raggedright
2.53
\end{minipage} \\
\begin{minipage}[t]{\linewidth}\raggedright
Average relative error (\%)
\end{minipage} & \begin{minipage}[t]{\linewidth}\raggedright
0.42
\end{minipage} & \begin{minipage}[t]{\linewidth}\raggedright
0.80
\end{minipage} & \begin{minipage}[t]{\linewidth}\raggedright
0.94
\end{minipage} & \begin{minipage}[t]{\linewidth}\raggedright
0.80
\end{minipage} & \begin{minipage}[t]{\linewidth}\raggedright
0.50
\end{minipage} & \begin{minipage}[t]{\linewidth}\raggedright
0.51
\end{minipage} \\
\begin{minipage}[t]{\linewidth}\raggedright
Average absolute error (\(\si{\celsius}\))
\end{minipage} & \begin{minipage}[t]{\linewidth}\raggedright
0.090
\end{minipage} & \begin{minipage}[t]{\linewidth}\raggedright
0.176
\end{minipage} & \begin{minipage}[t]{\linewidth}\raggedright
0.194
\end{minipage} & \begin{minipage}[t]{\linewidth}\raggedright
0.164
\end{minipage} & \begin{minipage}[t]{\linewidth}\raggedright
0.105
\end{minipage} & \begin{minipage}[t]{\linewidth}\raggedright
0.106
\end{minipage} \\
\begin{minipage}[t]{\linewidth}\raggedright
Standard deviation (\(\si{\celsius}\))
\end{minipage} & \begin{minipage}[t]{\linewidth}\raggedright
0.097
\end{minipage} & \begin{minipage}[t]{\linewidth}\raggedright
0.134
\end{minipage} & \begin{minipage}[t]{\linewidth}\raggedright
0.144
\end{minipage} & \begin{minipage}[t]{\linewidth}\raggedright
0.131
\end{minipage} & \begin{minipage}[t]{\linewidth}\raggedright
0.091
\end{minipage} & \begin{minipage}[t]{\linewidth}\raggedright
0.083
\end{minipage} \\
\end{longtable}

The Multiphysics simulation model enables rapid acquisition of indoor temperature field data under various scenarios and operating conditions. Compared to traditional experimental data collection methods, simulation-based data generation significantly reduces the complexity and time costs associated with data acquisition while greatly enhancing the diversity and coverage of the dataset. Therefore, the Multiphysics simulation model is employed to simulate multiple representative scenarios, generating high-resolution temperature field data. This synthetic data, combined with experimentally collected measurements, is used to construct the training and testing datasets for the PINN model, thereby improving its learning efficiency and generalization capability.

\subsection{PINN Model Training}

\begin{figure}[htbp]
\centering
\includegraphics[width=0.80\linewidth]{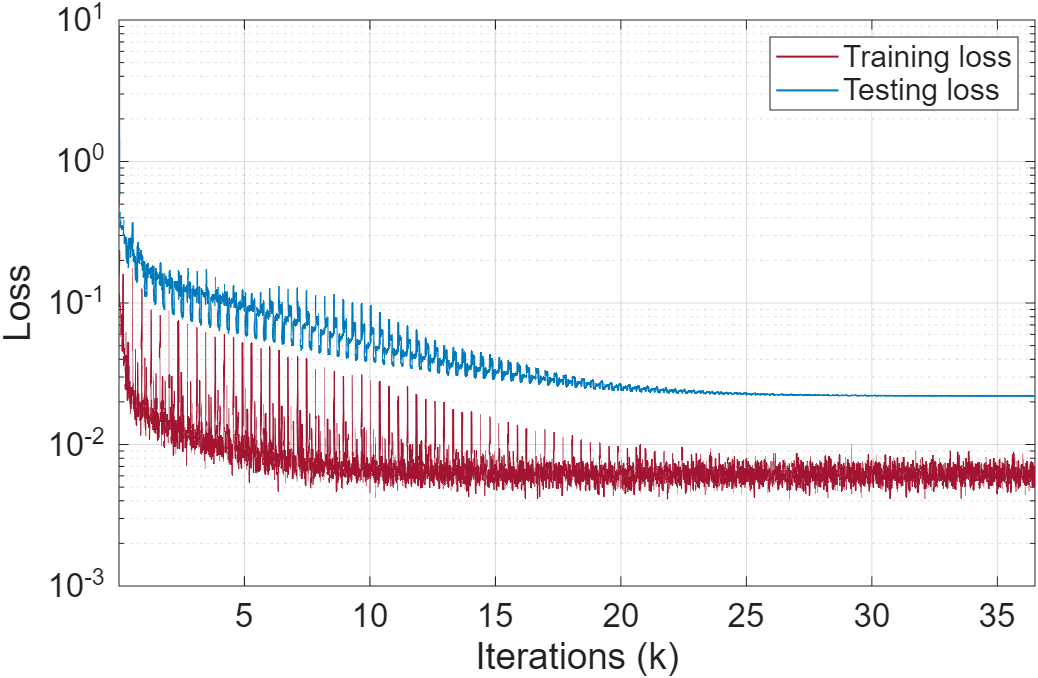}
\caption{Training loss versus iterations (1k=1000).}
\label{fig:5}
\end{figure}

Fig.~\ref{fig:5} shows the variation of training and testing losses with the number of iterations. Initially, both losses exhibit a sharp decline, indicating rapid convergence during the early stages of training. As the number of iterations increases, the training loss stabilizes at a low value, and the testing loss also reaches a stable value. This indicates that the PINN successfully balances fitting the data and adhering to the imposed physical constraints.

\begin{figure}[htbp]
\centering
\includegraphics[width=\linewidth,height=0.78\textheight,keepaspectratio]{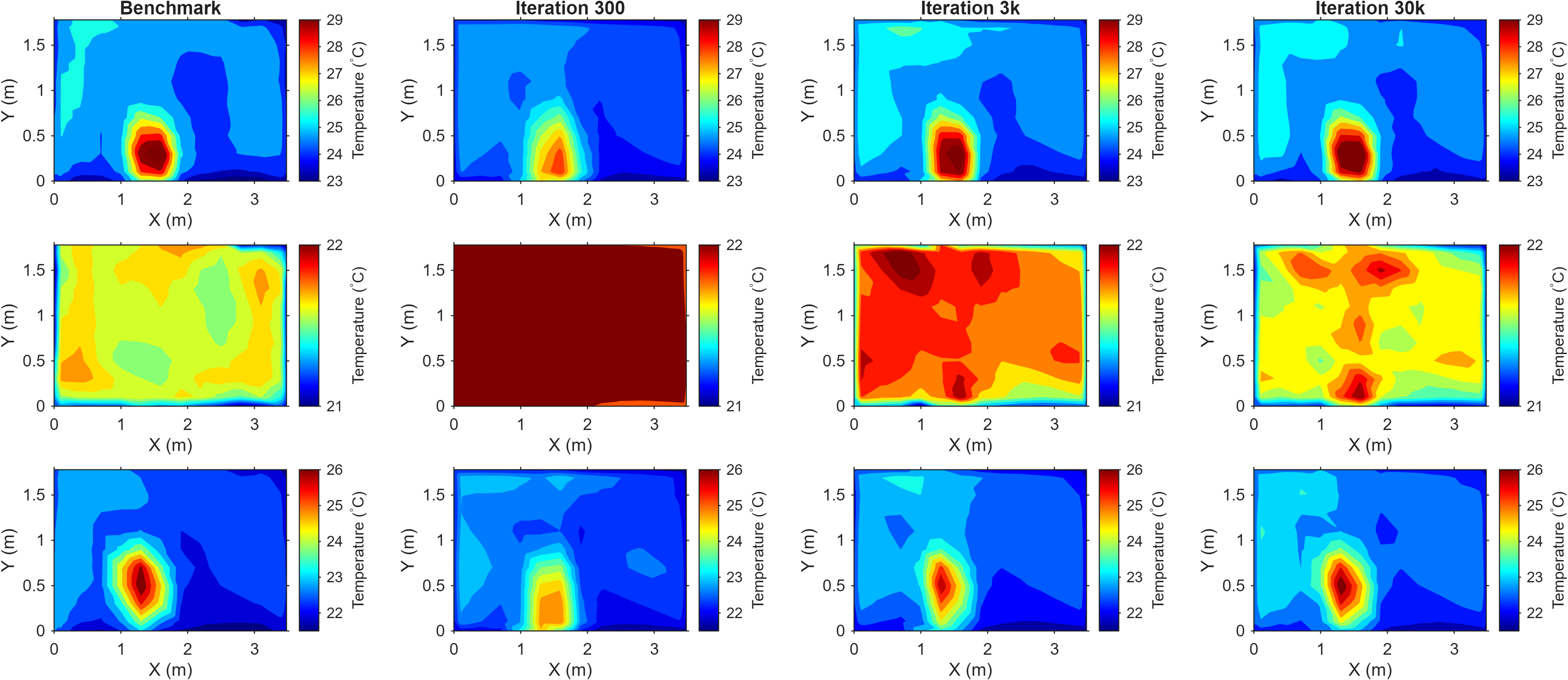}
\caption{Comparison of ground truth and predictions at different training iterations.}
\label{fig:6}
\end{figure}

Fig.~\ref{fig:6} presents a comparison between the actual temperature distribution of the target plane (first column) and the temperature fields reconstructed by the proposed PINN model at different training iterations: 300 (second column), 3,000 (third column), and 30,000 (fourth column). The temperature distributions on the target plane at three moments are selected as benchmarks to evaluate the performance of PINN.

At 300 iterations, the PINN model struggled to capture the temperature field, resulting in coarse and inaccurate predictions with minimal spatial detail. The mean squared errors (MSEs) across all spatial points for the three cases were 0.881, 0.432, and 1.255, respectively. When the number of iterations increased to 3,000, the predicted temperature field showed significant improvement. The primary temperature distribution patterns became more consistent with the reference solution, although discrepancies remained in finer details. Correspondingly, the MSEs decreased to 0.037, 0.368, and 0.060. At 30,000 iterations, the PINN model demonstrated strong performance, accurately capturing both the global temperature distribution and finer local variations. The MSEs were further reduced to 0.012, 0.178, and 0.018, indicating a substantial improvement in prediction accuracy.

These results indicate that the accuracy of the PINN model improves with the number of iterations, successfully reconstructing detailed temperature fields. This demonstrates the potential of the PINN model in accurately capturing the temperature field prediction. Therefore, the 36,000 iterations used in this study adequately support PINN's ability to learn the intricate details of the temperature field.

\subsection{PINN Model Validation}

\begin{figure}[htbp]
\centering
\includegraphics[width=0.80\linewidth]{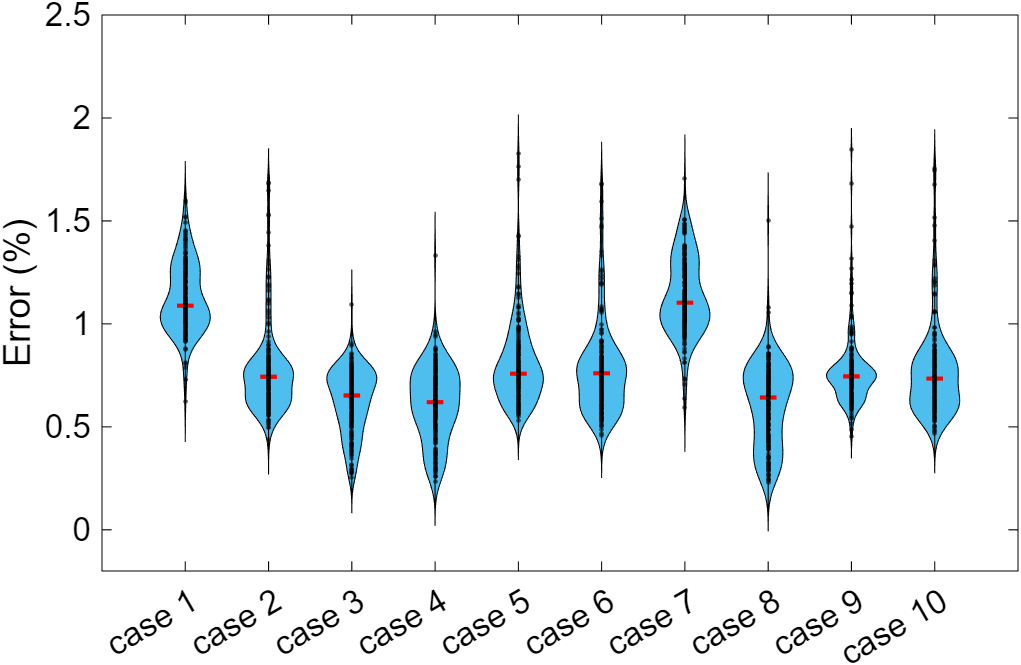}
\caption{Violin plots of relative error distributions for ten validation cases. Each violin shows n=180 validation time-step samples per case. Black points indicate individual validation samples. The violin width represents the kernel-density estimate of the error distribution. Red horizontal lines denote the median, and black vertical lines denote the interquartile range (25th-75th percentiles).}
\label{fig:7}
\end{figure}

Fig.~\ref{fig:7} presents the error distribution of the ten validation cases using violin plots. As shown in the figure, the error distribution for each case remains consistently low. The performance of the PINN on the remaining eight planes across the three room configurations is presented in the Supplementary Materials, where the errors are consistently maintained at a relatively low level.

The relative and absolute errors corresponding to each case are summarised in Table~\ref{tab:2}. Across all ten validation cases, the maximum relative error and average relative error of the predicted temperature fields at each time step were less than 1.85\% and 1.15\%, respectively. These results demonstrate the accuracy and robustness of the proposed PINN in predicting temperature fields across various validation cases.

\begingroup
\footnotesize
\setlength{\tabcolsep}{3pt}
\begin{longtable}[]{@{}
  >{\raggedright\arraybackslash}p{(\linewidth - 20\tabcolsep) * \real{0.1580}}
  >{\raggedright\arraybackslash}p{(\linewidth - 20\tabcolsep) * \real{0.0790}}
  >{\raggedright\arraybackslash}p{(\linewidth - 20\tabcolsep) * \real{0.0849}}
  >{\raggedright\arraybackslash}p{(\linewidth - 20\tabcolsep) * \real{0.0848}}
  >{\raggedright\arraybackslash}p{(\linewidth - 20\tabcolsep) * \real{0.0848}}
  >{\raggedright\arraybackslash}p{(\linewidth - 20\tabcolsep) * \real{0.0848}}
  >{\raggedright\arraybackslash}p{(\linewidth - 20\tabcolsep) * \real{0.0848}}
  >{\raggedright\arraybackslash}p{(\linewidth - 20\tabcolsep) * \real{0.0848}}
  >{\raggedright\arraybackslash}p{(\linewidth - 20\tabcolsep) * \real{0.0848}}
  >{\raggedright\arraybackslash}p{(\linewidth - 20\tabcolsep) * \real{0.0848}}
  >{\raggedright\arraybackslash}p{(\linewidth - 20\tabcolsep) * \real{0.0848}}@{}}
\caption{PINN validation errors across different cases.}\label{tab:2}\tabularnewline
\toprule\noalign{}
\begin{minipage}[b]{\linewidth}\raggedleft
Cases

Error
\end{minipage} & \begin{minipage}[b]{\linewidth}\raggedright
1
\end{minipage} & \begin{minipage}[b]{\linewidth}\raggedright
2
\end{minipage} & \begin{minipage}[b]{\linewidth}\raggedright
3
\end{minipage} & \begin{minipage}[b]{\linewidth}\raggedright
4
\end{minipage} & \begin{minipage}[b]{\linewidth}\raggedright
5
\end{minipage} & \begin{minipage}[b]{\linewidth}\raggedright
6
\end{minipage} & \begin{minipage}[b]{\linewidth}\raggedright
7
\end{minipage} & \begin{minipage}[b]{\linewidth}\raggedright
8
\end{minipage} & \begin{minipage}[b]{\linewidth}\raggedright
9
\end{minipage} & \begin{minipage}[b]{\linewidth}\raggedright
10
\end{minipage} \\
\midrule\noalign{}
\endfirsthead
\toprule\noalign{}
\begin{minipage}[b]{\linewidth}\raggedleft
Cases

Error
\end{minipage} & \begin{minipage}[b]{\linewidth}\raggedright
1
\end{minipage} & \begin{minipage}[b]{\linewidth}\raggedright
2
\end{minipage} & \begin{minipage}[b]{\linewidth}\raggedright
3
\end{minipage} & \begin{minipage}[b]{\linewidth}\raggedright
4
\end{minipage} & \begin{minipage}[b]{\linewidth}\raggedright
5
\end{minipage} & \begin{minipage}[b]{\linewidth}\raggedright
6
\end{minipage} & \begin{minipage}[b]{\linewidth}\raggedright
7
\end{minipage} & \begin{minipage}[b]{\linewidth}\raggedright
8
\end{minipage} & \begin{minipage}[b]{\linewidth}\raggedright
9
\end{minipage} & \begin{minipage}[b]{\linewidth}\raggedright
10
\end{minipage} \\
\midrule\noalign{}
\endhead
\bottomrule\noalign{}
\endlastfoot
\begin{minipage}[t]{\linewidth}\raggedright
Maximum relative error (\%)
\end{minipage} & \begin{minipage}[t]{\linewidth}\raggedright
1.60
\end{minipage} & \begin{minipage}[t]{\linewidth}\raggedright
1.69
\end{minipage} & \begin{minipage}[t]{\linewidth}\raggedright
1.09
\end{minipage} & \begin{minipage}[t]{\linewidth}\raggedright
1.33
\end{minipage} & \begin{minipage}[t]{\linewidth}\raggedright
1.83
\end{minipage} & \begin{minipage}[t]{\linewidth}\raggedright
1.68
\end{minipage} & \begin{minipage}[t]{\linewidth}\raggedright
1.71
\end{minipage} & \begin{minipage}[t]{\linewidth}\raggedright
1.50
\end{minipage} & \begin{minipage}[t]{\linewidth}\raggedright
1.84
\end{minipage} & \begin{minipage}[t]{\linewidth}\raggedright
1.75
\end{minipage} \\
\begin{minipage}[t]{\linewidth}\raggedright
Average relative error (\%)
\end{minipage} & \begin{minipage}[t]{\linewidth}\raggedright
1.13
\end{minipage} & \begin{minipage}[t]{\linewidth}\raggedright
0.78
\end{minipage} & \begin{minipage}[t]{\linewidth}\raggedright
0.63
\end{minipage} & \begin{minipage}[t]{\linewidth}\raggedright
0.61
\end{minipage} & \begin{minipage}[t]{\linewidth}\raggedright
0.81
\end{minipage} & \begin{minipage}[t]{\linewidth}\raggedright
0.78
\end{minipage} & \begin{minipage}[t]{\linewidth}\raggedright
1.14
\end{minipage} & \begin{minipage}[t]{\linewidth}\raggedright
0.61
\end{minipage} & \begin{minipage}[t]{\linewidth}\raggedright
0.77
\end{minipage} & \begin{minipage}[t]{\linewidth}\raggedright
0.77
\end{minipage} \\
\begin{minipage}[t]{\linewidth}\raggedright
Average absolute error (\(\si{\celsius}\))
\end{minipage} & \begin{minipage}[t]{\linewidth}\raggedright
0.243
\end{minipage} & \begin{minipage}[t]{\linewidth}\raggedright
0.166
\end{minipage} & \begin{minipage}[t]{\linewidth}\raggedright
0.135
\end{minipage} & \begin{minipage}[t]{\linewidth}\raggedright
0.133
\end{minipage} & \begin{minipage}[t]{\linewidth}\raggedright
0.177
\end{minipage} & \begin{minipage}[t]{\linewidth}\raggedright
0.170
\end{minipage} & \begin{minipage}[t]{\linewidth}\raggedright
0.243
\end{minipage} & \begin{minipage}[t]{\linewidth}\raggedright
0.131
\end{minipage} & \begin{minipage}[t]{\linewidth}\raggedright
0.165
\end{minipage} & \begin{minipage}[t]{\linewidth}\raggedright
0.162
\end{minipage} \\
\begin{minipage}[t]{\linewidth}\raggedright
Standard deviation (\(\si{\celsius}\))
\end{minipage} & \begin{minipage}[t]{\linewidth}\raggedright
0.043
\end{minipage} & \begin{minipage}[t]{\linewidth}\raggedright
0.053
\end{minipage} & \begin{minipage}[t]{\linewidth}\raggedright
0.037
\end{minipage} & \begin{minipage}[t]{\linewidth}\raggedright
0.042
\end{minipage} & \begin{minipage}[t]{\linewidth}\raggedright
0.051
\end{minipage} & \begin{minipage}[t]{\linewidth}\raggedright
0.057
\end{minipage} & \begin{minipage}[t]{\linewidth}\raggedright
0.044
\end{minipage} & \begin{minipage}[t]{\linewidth}\raggedright
0.046
\end{minipage} & \begin{minipage}[t]{\linewidth}\raggedright
0.041
\end{minipage} & \begin{minipage}[t]{\linewidth}\raggedright
0.050
\end{minipage} \\
\end{longtable}
\endgroup

\begin{figure}[htbp]
\centering
\includegraphics[width=\linewidth,height=0.78\textheight,keepaspectratio]{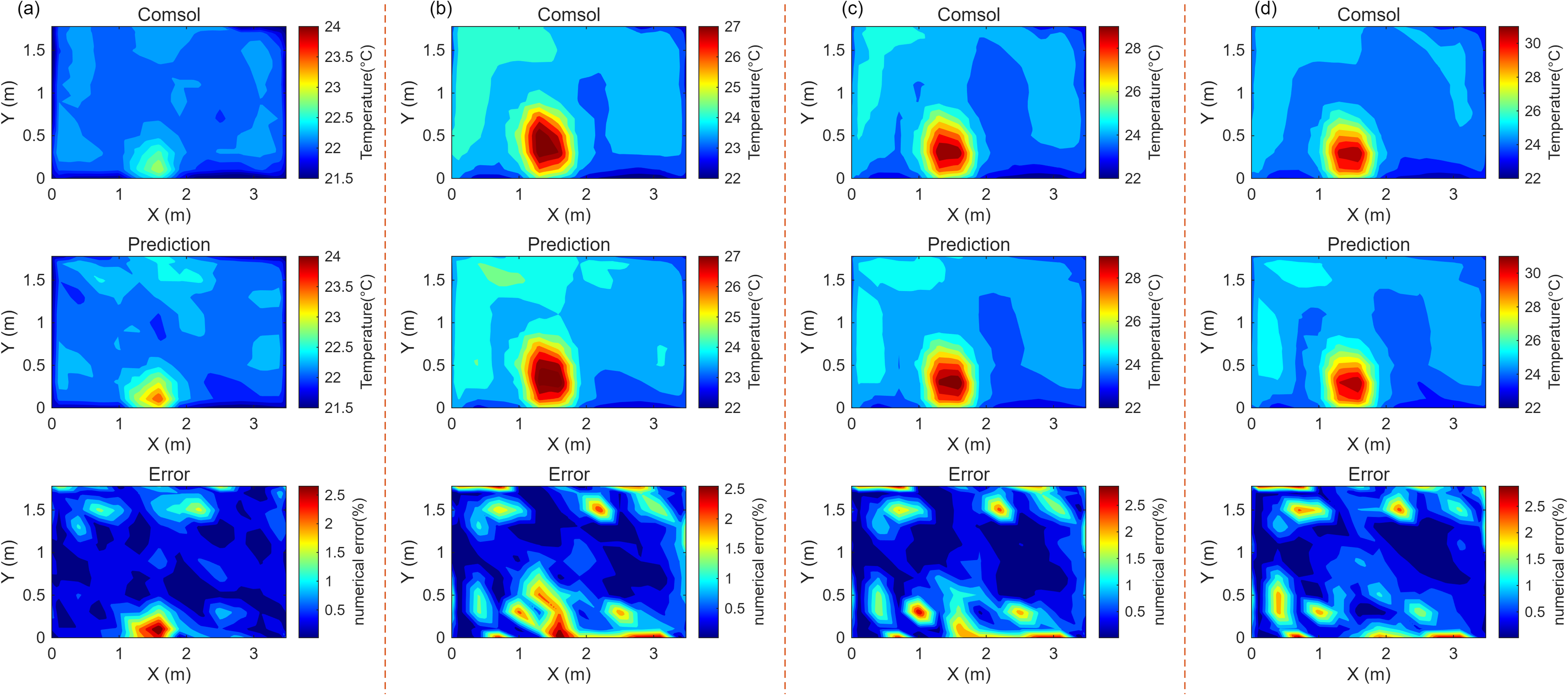}
\caption{Temperature inversion performance of the trained PINN under different heating scenarios. (a) Heater-off mode. (b) Heater operating at 1 kW. (c) Heater operating at 2 kW. (d) Heater operating at 3 kW.}
\label{fig:8}
\end{figure}

In each test case, the heater operating mode is randomly sampled at each simulation time step. To provide a more detailed analysis, Case 5 is selected as a representative example. Four scenarios corresponding to different heater operating modes are considered: heater-off mode, and heater operating at 1 kW, 2 kW, and 3 kW. Fig.~\ref{fig:8} compares the COMSOL simulation results with the predicted temperature fields obtained from the trained PINN model for these four heating scenarios, along with the errors between the two.

The results indicate that the trained PINN model can inverse the temperature fields under these four heating scenarios with acceptable accuracy. The error distributions across different heater operating modes are consistently small. The reconstructed temperature fields from the PINN model closely agree with the COMSOL simulation results, capturing both the global temperature distribution characteristics and the localized high- and low-temperature regions. This validates the effectiveness and reliability of the proposed PINN in predicting temperature fields under varying heating conditions.

\subsection{Occupant-Centric Space Heating Control Performance}

\begin{figure}[htbp]
\centering
\includegraphics[width=\linewidth,height=0.78\textheight,keepaspectratio]{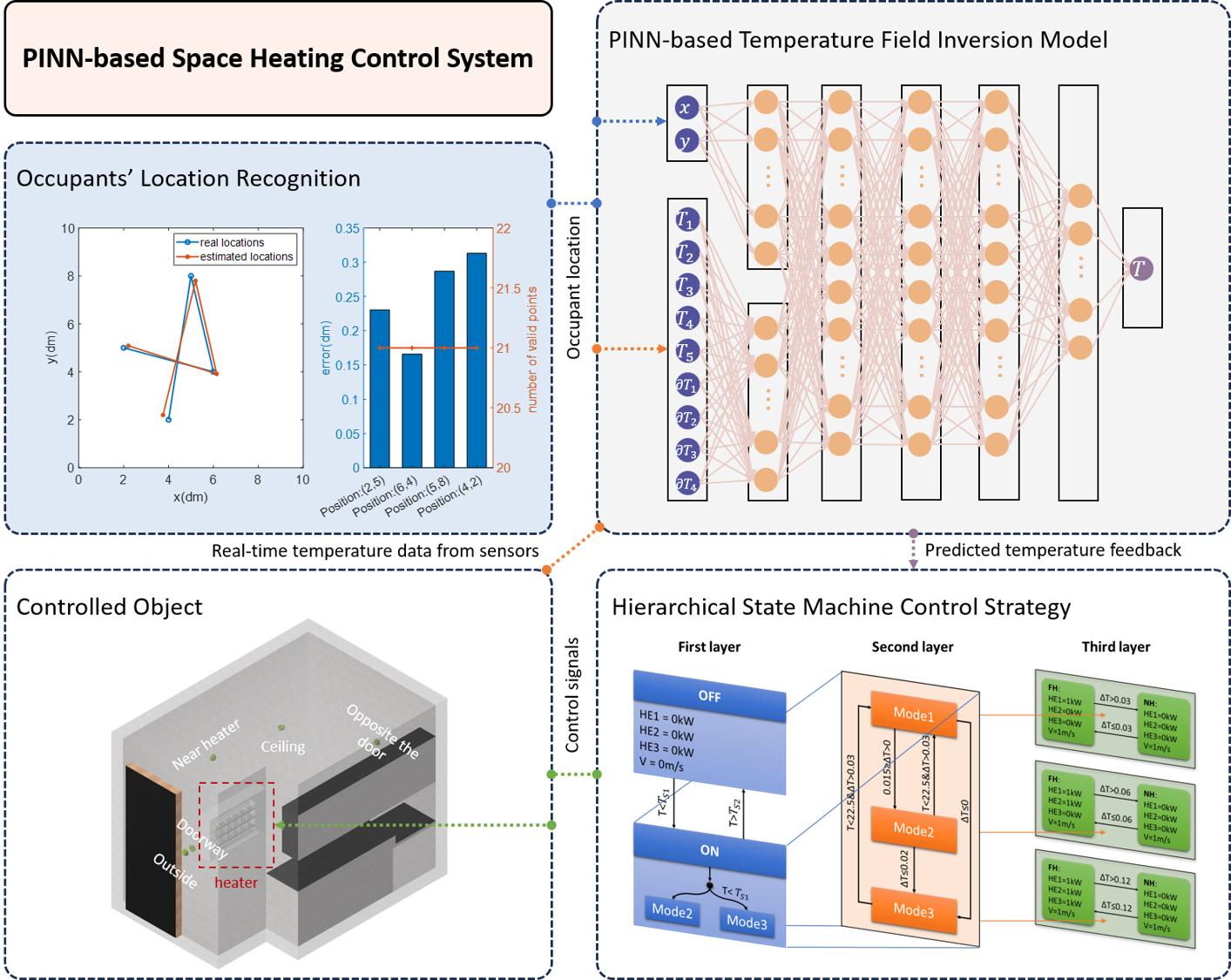}
\caption{Framework of the PINN-based space heating control system.}
\label{fig:9}
\end{figure}

Figure~\ref{fig:9} summarises the evaluated PINN-based space heating control framework. The COMSOL-based implementation, including its hierarchical state-machine controller, occupant-location input, and benchmark performance, was reported in our previous studies \cite{main4,main58}. In that benchmark, the COMSOL-based controller improved thermal comfort by up to 38\% while reducing energy consumption by approximately 20\% relative to two conventional control systems. Its reliance on computationally intensive commercial Multiphysics simulation, however, limits real-time responsiveness. The present comparison reuses the controller and matched test scenario while replacing only the COMSOL temperature-field estimator with the trained PINN, allowing the estimator accuracy and computation time to be compared directly.

On the computer used for this study, the PINN-based space heating control system and the COMSOL-based space heating control system required 60.69 seconds (approximately 1 minute) and 31,923.91 seconds (approximately 8.8 hours), respectively, to complete a 3-hour simulation case. The former system completed control decision-making in only 0.56\% of real-time, whereas the latter required 2.93 times the real-time to finalize decisions. This significant delay in the COMSOL-based system would adversely affect control performance during practical applications. Furthermore, the processing capability of actual controllers is significantly lower than that of the computer used in this simulation, which would further increase the decision-making time. Consequently, the COMSOL-based space heating control system is challenging to implement in real-world scenarios. In contrast, the PINN-based space heating control system demonstrates much faster computation, supporting its potential for real-time implementation; deployment on lower-performance controllers remains to be validated.

\begin{figure}[htbp]
\centering
\includegraphics[width=0.90\linewidth]{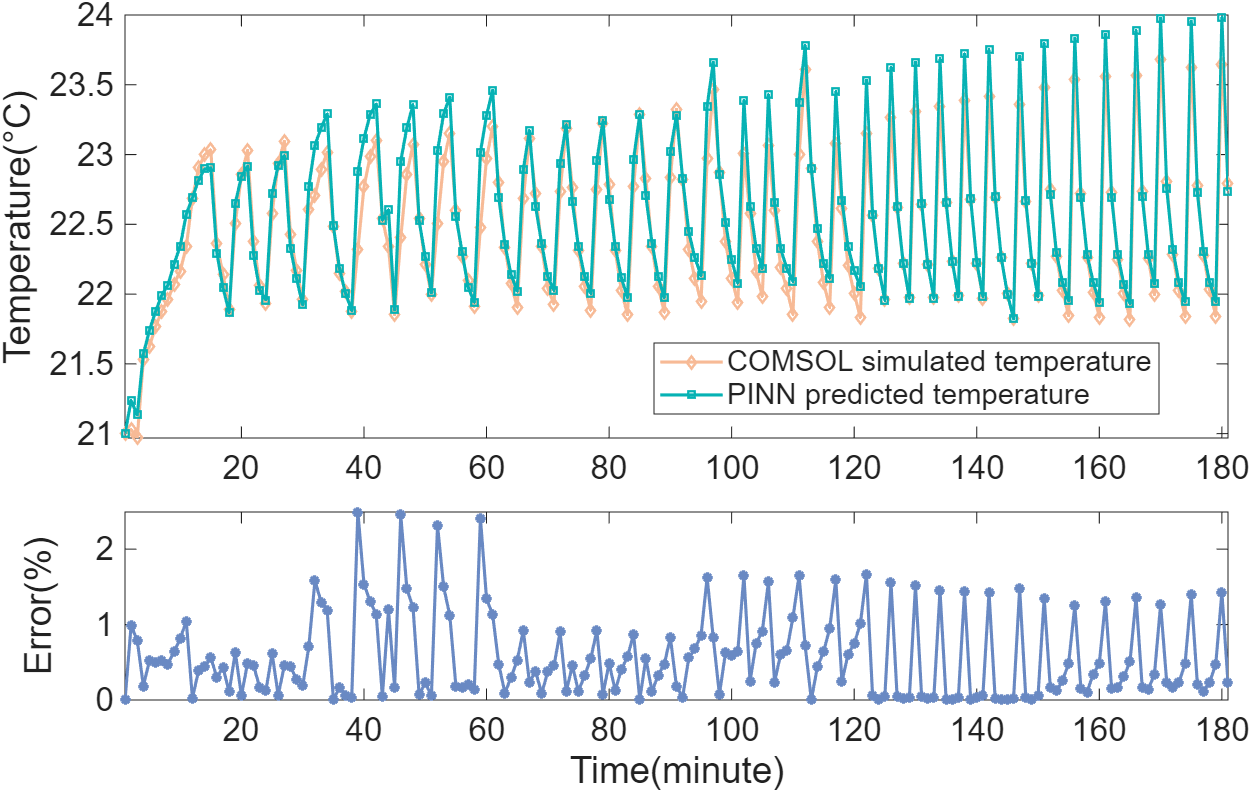}
\caption{Comparison of occupant-surrounding temperature between COMSOL simulations and PINN predictions during control process.}
\label{fig:10}
\end{figure}

At the end of the simulation, the temperatures along the occupant\textquotesingle s activity trajectory were extracted, and the results are presented in Fig.~\ref{fig:10}. Both control systems successfully maintained the temperature around the occupant within the target range. The differences between the temperatures predicted by the COMSOL simulation and those by the PINN model were minimal. The numerical errors between the two temperature datasets were calculated using Eq.~\eqref{eq:5}, and the time-varying numerical errors are detailed in Fig.~\ref{fig:10}. The maximum numerical error between these datasets was 2.46\%, and the average numerical error was 0.67\%.

The results demonstrate that the PINN-based space heating control system achieves performance comparable to that of the COMSOL-based control system, while significantly reducing computational complexity. It maintains accurate temperature regulation and achieves similar levels of occupant thermal comfort and energy efficiency under the evaluated control scenario, while overcoming the limitations of high computational cost and long simulation times. This highlights its potential for real-time occupant-centric control in comparable room-scale heating applications.

\section{Conclusions}

This study presents a control-oriented Physics-Informed Neural Network (PINN) architecture for space heating control systems. By eliminating the need for explicit time inputs and instead leveraging sensor feedback to infer temporal dynamics, the proposed method addresses key limitations of traditional PINN structures. The model is trained using a combination of experimental and simulation data, achieving accurate reconstructions of dynamic temperature fields with reduced computational cost.

The robustness and accuracy of the model were validated against COMSOL Multiphysics benchmarks, with the maximum error remaining below 1.85\%. When integrated into the space heating control system, the PINN-based space heating control system demonstrated an average numerical error of only 0.67\%, while reducing computational time to 0.19\% of that required by the COMSOL-based system, highlighting its suitability for real-time control applications. Furthermore, the system effectively maintained occupant thermal comfort while simultaneously reducing energy consumption.

Practical deployment challenges, such as sensor failures and data reliability issues, will be systematically analysed, together with corresponding mitigation strategies. Future work will also explore integration of the proposed framework with more advanced control strategies, such as Model Predictive Control (MPC) and Reinforcement Learning (RL), to further enhance control performance and adaptability.

\section*{Acknowledgements}

J.W. discloses support for the research of this work from the UK Engineering and Physical Sciences Research Council (EPSRC; grant numbers EP/S032622/1 and EP/W027372/1).

\section*{Data and code availability}

The data and code are available in the following GitHub repository:\\
{\urlstyle{same}\url{https://github.com/DanialBai/control-oriented-pinn}}.

\clearpage

\end{document}